\documentclass[letterpaper]{article} 
\usepackage{aaai25}  
\usepackage{times}  
\usepackage{helvet}  
\usepackage{courier}  
\usepackage[hyphens]{url}  
\usepackage{graphicx} 
\usepackage{natbib}  
\usepackage{caption} 
\usepackage{algorithm}
\usepackage{algorithmic}
\usepackage{amsmath}
\usepackage{booktabs}
\usepackage{multirow}

\usepackage{newfloat}
\usepackage{listings}
\DeclareCaptionStyle{ruled}{labelfont=normalfont,labelsep=colon,strut=off} 
\floatstyle{ruled}
\newfloat{listing}{tb}{lst}{}
\floatname{listing}{Listing}
\title{Video Summarization using Denoising Diffusion Probabilistic Model}
\author{
    Zirui Shang\textsuperscript{\rm 1},
    Yubo Zhu\textsuperscript{\rm 1},
    Hongxi Li\textsuperscript{\rm 1},
    Shuo Yang\textsuperscript{\rm 2},
    Xinxiao Wu\textsuperscript{\rm 1, \rm 2}\thanks{Xinxiao Wu is the corresponding author.}
}
\affiliations{
    \textsuperscript{\rm 1}Beijing Key Laboratory of Intelligent Information Technology, School of Computer Science \& Technology,\\ Beijing Institute of Technology, China\\
    \textsuperscript{\rm 2}Guangdong Laboratory of Machine Perception and Intelligent Computing, Shenzhen MSU-BIT University, China\\
    \{shangzirui, 3120211052, lihongxi, wuxinxiao\}@bit.edu.cn, yangshuo@smbu.edu.cn
}

\begin{document}

\maketitle

\begin{abstract}
Video summarization aims to eliminate visual redundancy while retaining key parts of video to construct concise and comprehensive synopses. Most existing methods use discriminative models to predict the importance scores of video frames. However, these methods are susceptible to annotation inconsistency caused by the inherent subjectivity of different annotators when annotating the same video. In this paper, we introduce a generative framework for video summarization that learns how to generate summaries from a probability distribution perspective, effectively reducing the interference of subjective annotation noise. Specifically, we propose a novel diffusion summarization method based on the Denoising Diffusion Probabilistic Model (DDPM), which learns the probability distribution of training data through noise prediction, and generates summaries by iterative denoising. Our method is more resistant to subjective annotation noise, and is less prone to overfitting the training data than discriminative methods, with strong generalization ability. Moreover, to facilitate training DDPM with limited data, we employ an unsupervised video summarization model to implement the earlier denoising process. Extensive experiments on various datasets (TVSum, SumMe, and FPVSum) demonstrate the effectiveness of our method.
\end{abstract}

%

\section{Introduction}

With the popularity of video-sharing platforms and social media, video data is experiencing explosive growth, and increasing attentions are paid to automatically identifying and extracting representative segments from a video. Video summarization emerges as a critical technique that condenses video content into a concise summary while preserving essential information and key moments. These summaries enable users to quickly grasp the video content, making it more easily accessible and manageable in various scenarios, such as online platforms, surveillance systems, and multimedia archives.

\begin{figure}[t]
  \centering
  \includegraphics[width=\linewidth]{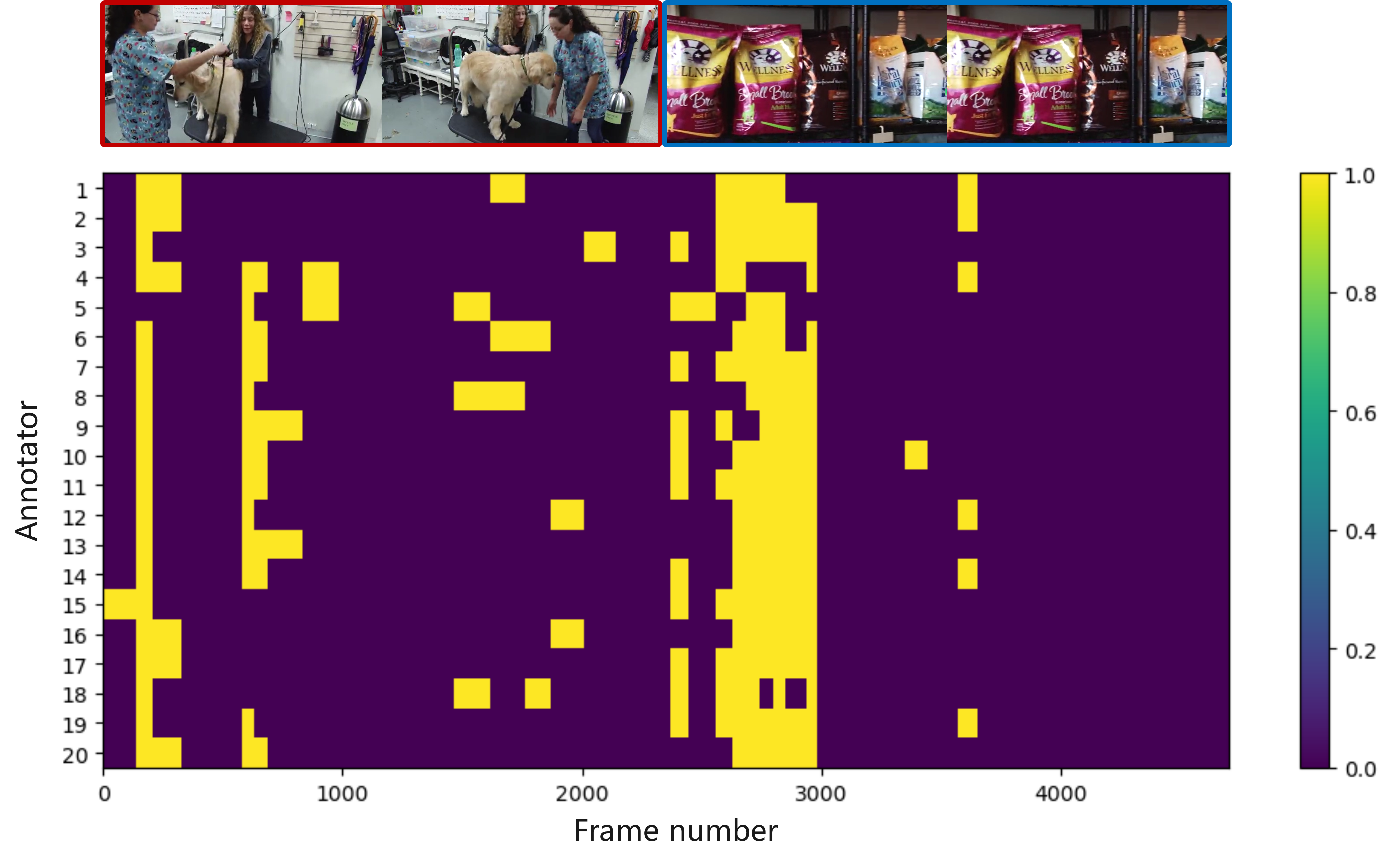}
  \caption{An example of subjective annotation noise in the TVSum dataset, where yellow blocks represent the annotated video frames of summaries by different annotators.}
  \label{Figure1}
\end{figure}

Considerable progress has been made in video summarization recently, and many deep models have been applied, such as Long Short Term Memory (LSTM)~\cite{zhang2016video}, Fully Convolutional Network (FCN)~\cite{liang2022video}, Transformer~\cite{li2023progressive, hsu2023video, terbouche2023multi}, etc. However, these existing methods use discriminative models to predict the importance scores of video frames, which are susceptible to annotation inconsistency. The inconsistency is inevitable because of the inherent subjectivity of different annotators when annotating the same video. For example, as shown in Figure~\ref{Figure1}, when facing the video of a pet store advertisement, some annotators may focus on the cute pets in the video (shown in the red box), and others may show more interest in the products in the store (shown in the blue box), resulting in different annotations of importance scores.

To address this challenge, we introduce a generative framework for video summarization that learns how to generate summaries from a probability distribution perspective, effectively reducing the interference of subjective annotation noise. In this paper, we propose a novel diffusion summarization method based on the Denoising Diffusion Probabilistic Model (DDPM), which learns the probability distribution of training data through noise prediction, and generates summaries by iterative denoising. Specifically, we use video frame features as guidance, and take the importance scores after adding noise as input to train DDPM for generating clear importance scores. We directly train DDPM using raw annotations rather than averaging them, enabling it to learn the distribution of training data. Therefore, our method is more resistant to subjective annotation noise and simultaneously less prone to overfitting the training set, with stronger generalization ability than existing discriminative methods.

Moreover, we observe that DDPM faces performance degradation with limited training data, and combine an unsupervised video summarization model with it to address this issue. During training, we reduce the number of maximum noise addition steps, allowing DDPM to start denoising from a relatively clear intermediate result instead of a Gaussian noise.
During testing, we use the output of the unsupervised video summarization model as the intermediate result mentioned above, providing an alternative to early denoising and achieving superior performance.
This strategy provides a feasible solution for DDPM training under data scarcity.

The main contributions of our work are summarized in three folds:
(1) We introduce a generative framework for video summarization, which learns to generate summaries by modeling the probability distribution of training data, effectively reducing the interference of subjective annotation noise.
(2) We propose a novel diffusion summarization method based on DDPM, which generates summaries by iterative denoising, and combines it with an unsupervised video summarization model to facilitate training with limited data.
(3) The effectiveness of our method is demonstrated through comprehensive experiments on three benchmark datasets: TVSum, SumMe, and FPVSum.


\section{Related Work}
\subsection{Video Summarization}
Early traditional methods of video summarization rely on low-level visual features to extract key frames, such as color histogram~\cite{de2011vsumm}, spatiotemporal feature~\cite{laganiere2008video}, and motion cues~\cite{ren2017unsupervised}. With the considerable progress of deep learning in video processing and understanding, video summarization is formulated as an importance score prediction problem. Zhang et al.~\cite{zhang2016video} use LSTM to predict importance scores by modeling the variable-range temporal dependency among video frames. Zhao et al.~\cite{zhao2017hierarchical} propose a hierarchical LSTM tailored to long video scenarios. More recently, the Graph Neural Networks (GNNs), Convolutional Neural Networks (CNNs), and the attention mechanism have been employed in video summarization. Zhong et al.~\cite{zhong2023semantic} propose a Contextual Feature Transformation (CFT) mechanism that builds upon Graph Information Bottle (GIB) to enhance the temporal correlation among images. Hsu et al.~\cite{hsu2023video} propose a novel transformer-based method named spatiotemporal vision transformer (STVT) for video summarization, which considers both of inter-frame correlations among non-adjacent frames and intra-frame attention which attracts humans. MPFN~\cite{khan2024deep} proposes a deep pyramidal refinement network to extract and refine multi-scale progressive features. 
LDH-based Deep CNN~\cite{singh2024bayesian} selects representative frames using Bayesian fuzzy clustering (BFC) and refines those frames using deep CNN.

All these methods rely on predicting the importance score of each frame from a discriminative perspective to generate summaries, which may be affected by subjective annotation noise. To reduce this interference, we attempt to generate summaries from a probability distribution perspective and introduce a generative framework for video summarization. Unlike existing generative models~\cite{he2019unsupervised, mahasseni2017unsupervised, apostolidis2020unsupervised} commonly used for unsupervised video summarization, our method introduces DDPM into supervised video summarization, which progressively pinpoints important content in the video and can predict more precise importance scores~\cite{li2023progressive}.

\subsection{Denoising Diffusion Probabilistic Model}
DDPM belongs to a category of generative models, which has emerged as super performers in producing high-quality samples. It is initially used in the field of image generation. Inspired by considerations from nonequilibrium thermodynamics, Ho et al.~\cite{ho2020denoising} present high-quality image synthesis results by using a diffusion probabilistic model. Subsequently, Nichol et al.~\cite{nichol2021improved} show that with a few simple modifications, DDPM can sample much faster and achieve better log-likelihoods with little impact on sample quality. Recently, DDPM has also been widely used in various fields, including image colorization~\cite{carrillo2023diffusart}, super-resolution~\cite{li2022srdiff, gao2023implicit}, image editing~\cite{wu2023latent}, and semantic segmentation~\cite{wu2023latent}.

We make the first attempt to introduce DDPM into the field of video summarization and use video frame features as guidance to recover the corresponding frame-level importance scores from noise. Extensive experiments on various datasets demonstrate that DDPM can be successfully used in video summarization.

\begin{figure*}[t]
  \centering
  \includegraphics[width=\textwidth]{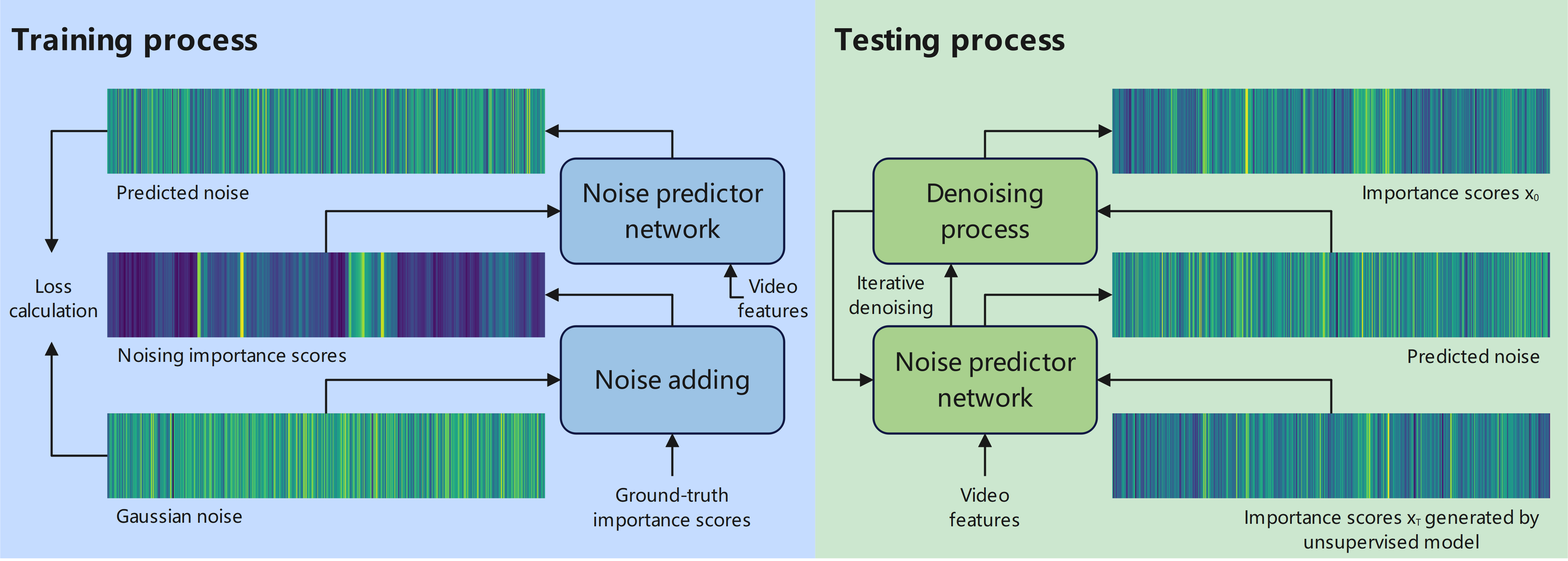}
  \caption{The framework of our method, where the training process shows how the noise predictor network learns to predict noise components, and the testing process shows how to generate accurate importance scores through denoising.}
  \label{Figure2}
\end{figure*}

\begin{figure}[t]
  \centering
  \includegraphics[width=0.8\linewidth]{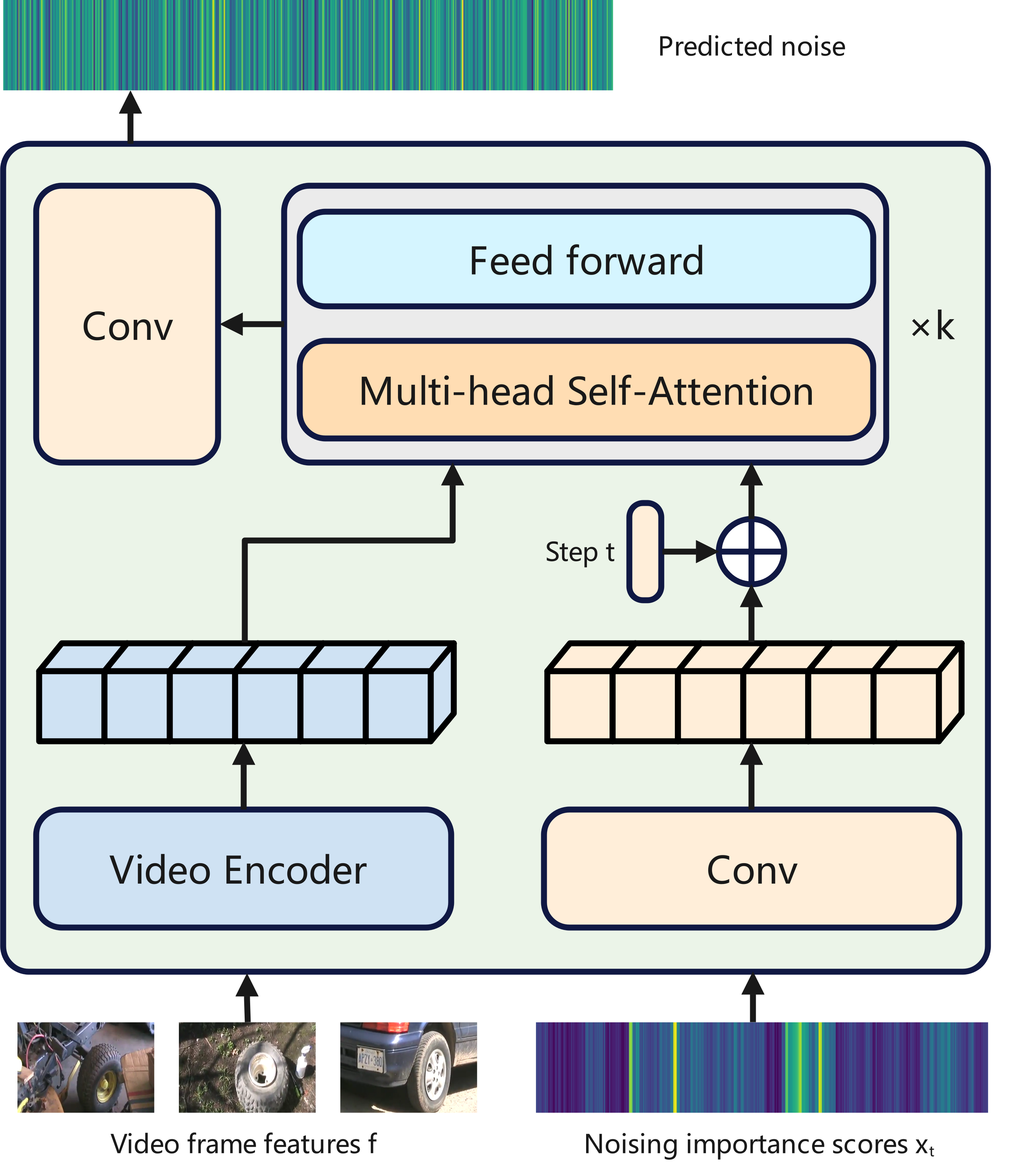}
  \caption{The structure of noise predictor network, which uses video frame features $f$ as guidance and noising importance scores $x_t$ as input to predict the noise component at step $t$.}
  \label{Figure3}
\end{figure}

\section{Our Method}
\subsection{Background}
In this section, we provide a brief overview of diffusion models. Diffusion models aim to convert the noise $x_t$ sampled from $\mathcal{N}(\textbf{0}, \textbf{I})$ into the desired output $x_0$, under the assuming that $x_t$ is equivalent to adding $t$ steps Gaussian noise to $x_0$, with fixed variance schedule $\beta_1,..., \beta_t$. Formally, this is defined as a forward diffusion process:
\begin{equation}
\begin{aligned}
x_t=\sqrt{1-\beta_t}x_{t-1}+\sqrt{\beta_t}\epsilon, \epsilon\sim \mathcal{N}(\textbf{0}, \textbf{I}), \\
q(x_t|x_{t-1})=\mathcal{N}(x_t;\sqrt{1-\beta_t}x_{t-1}, \beta_t\textbf{I}),
\end{aligned}
\end{equation}
where $q(x_t|x_{t-1})$ is the distribution of $x_t$ under the condition of given $x_{t-1}$. Importantly, a noisy sample $x_t$ can be obtained directly from the $x_0$:
\begin{equation}
\begin{aligned}
x_t=\sqrt{\bar{\alpha}_t}x_0+\sqrt{1-\bar{\alpha}_t}\epsilon, \epsilon\sim \mathcal{N}(\textbf{0}, \textbf{I}), \\
q(x_t|x_0)=\mathcal{N}(x_t;\sqrt{\bar{\alpha}_t}x_0, (1-\bar{\alpha}_t)\textbf{I}),
\end{aligned}
\end{equation}
where $\alpha_t=1-\beta_t$, $\bar{\alpha}_t=\prod_{s=1}^{t}\alpha_s$, and $q(x_t|x_0)$ is the distribution of $x_t$ under the condition of given $x_0$.

In particular, the diffusion model reverses the noising process, sampling from a distribution by gradually denoising, which starts with noise $x_t$ and learns to produce a slightly more ``denoised" $x_{t-1}$ from $x_t$, until reaching the final sample $x_0$. Using Bayes theorem, Ho et al.~\cite{ho2020denoising} define a forward process posteriors, which are tractable when conditioned on $x_0$:
\begin{equation}
\begin{aligned}
q(x_{t-1}|x_t, x_0)=\mathcal{N}(x_{t-1};\tilde{\mu}_t(x_t, x_0), \tilde{\beta}_t\textbf{I}),
\end{aligned}
\end{equation}
where $\tilde{\mu}_t(x_t, x_0)=\frac{\sqrt{\bar{\alpha}_{t-1}}\beta_t}{1-\bar{\alpha}_t}x_0+\frac{\sqrt{\alpha_t}(1-\bar{\alpha}_{t-1})}{1-\bar{\alpha}_t}x_t$, $\tilde{\beta}_t=\frac{1-\bar{\alpha_{t-1}}}{1-\bar{\alpha}_t}\beta_t$, and $q(x_{t-1}|x_t, x_0)$ is the distribution of $x_{t-1}$ under the condition of given $x_t$ and $x_0$. Unlike the $x_t$ sampled from $\mathcal{N}(\textbf{0}, \textbf{I})$, $x_0$ is unknown and cannot be obtained through sampling. Thus, to obtain the exact distribution $q(x_{t-1}|x_t)$, and run the forward diffusion process in reverse to obtain a distribution of $x_0$, a neural network is used to approximate $q(x_{t-1}|x_t)$ as follows:
\begin{equation}
\begin{aligned}
p_\theta(x_{t-1}|x_t)=\mathcal{N}(x_{t-1};\mu_\theta(x_t, t), \Sigma_\theta(x_t, t)), \\
\mu_\theta(x_t, t)=\frac{1}{\sqrt{\alpha_t}}(x_t-\frac{\beta_t}{\sqrt{1-\bar{\alpha}_t}}\epsilon_\theta(x_t, t)),
\end{aligned}
\end{equation}
where $\epsilon_\theta(x_t, t)$ is a noise predictor network to predict the noise component at the step $t$, $\Sigma_\theta(x_t, t)$ is a covariance predictor can be either a fixed set of scalar covariances or learned as well, and $p_\theta(x_{t-1}|x_t)$ is the approximate distribution of $x_{t-1}$ under the condition of given $x_t$. The noise predictor network $\epsilon_\theta(x_t, t)$ is usually selected from different variants of the UNet~\cite{ronneberger2015u} architecture, and with a simple mean-squared error loss.

\subsection{DDPM-based Video Summarization}
In the following sections, we describe our diffusion summarization method from two stages: the training process, which involves how the noise predictor network learns to predict noise components, and the testing process, which involves how to generate accurate importance scores through denoising. Figure~\ref{Figure2} illustrates the framework of our method.
\subsubsection{Training Process}
In the training process, we consider how the noise predictor network learns to predict noise components. The noise predictor network is defined as $\epsilon_\theta(x_t, f, t)$, which uses video frame features $f$ as guidance and noising importance scores $x_t$ as input to predict the noise component at step $t$. The structure of noise predictor network is shown in Figure~\ref{Figure3}. We use Transformer layers to build a bridge between importance scores and video frame features through attention mechanism. The importance scores are used as query vectors and the video frame features are used as key and value vectors to guide the model in predicting noise components. In addition, we linearly scale the ground-truth importance scores to [-1, 1] before adding noise. This ensures that the neural network reverse process operates on consistently scaled inputs starting the noise predictor network standard normal prior~\cite{ho2020denoising}. Then a complete noise prediction in the training process can be defined as follows:
\begin{equation}
\begin{aligned}
x_0&=Scale(x_g), \\
x_t&=\sqrt{\bar{\alpha}_t}x_0+\sqrt{1-\bar{\alpha}_t}\epsilon, \epsilon\sim \mathcal{N}(\textbf{0}, \textbf{I}), \\
\hat{\epsilon}&=\epsilon_\theta(x_t, f, t),
\label{predict}
\end{aligned}
\end{equation}
where $x_g$ represents the ground-truth importance scores, $Scale(\cdot)$ represents the linearly scaling to [-1, 1], $x_0$ represents the importance scores after scaling, $x_t$ represents the noising importance scores and $\hat{\epsilon}$ represents the predicted noise component.

The training objective is defined as a simple mean-squared error loss between the true noise and the predicted noise as follows:
\begin{equation}
\begin{aligned}
L=\left\|\epsilon-\hat{\epsilon}\right\|^2,
\label{loss}
\end{aligned}
\end{equation}
where $\epsilon$ represents the true noise and $L$ represents the loss. Algorithm 1 displays the complete training process, with $M$ epochs, $N$ videos, and maximum noise addition steps $T$. For each video, we traverse the importance scores annotated by different annotators, input it into the noise predictor network along with the corresponding video features, and optimize parameters through the loss function. 
\begin{algorithm}[tb]
\caption{Training Process}
\textbf{Input:} Video features with annotated importance scores. \\
\textbf{Output:} Parameters $\theta$ of the noise predictor network.
\begin{algorithmic}
\FOR {$epoch\in\{1, 2, ..., M\}$}
    \FOR {$n\in\{1, 2, ..., N\}$}
        \STATE Extract video features $f$ of the $n$-th video.
        \FOR {every importance scores annotation $x_g$}
            \STATE $t\sim$ Uniform$(\{1, ..., T\})$.
            \STATE $\epsilon\sim \mathcal{N}(\textbf{0}, \textbf{I})$.
            \STATE Predict noise $\hat{\epsilon}$ using Eq.~\ref{predict}.
            \STATE Calculate the loss $L$ using Eq.~\ref{loss}.
            \STATE Optimize the parameters $\theta$: $\theta \gets \theta-\eta~\partial L/\partial \theta$.
        \ENDFOR
    \ENDFOR
\ENDFOR
\end{algorithmic}
\end{algorithm}

\subsubsection{Testing Process}
In the testing process, we consider how to generate accurate importance scores through denoising. We observe that DDPM faces performance degradation with limited training data. To address this issue, we combine an unsupervised video summarization model~\cite{zhou2018deep} with DDPM, using its output as the start of denoising, providing an alternative to early denoising of DDPM. In addition, before denoising, we linearly scale the importance scores generated by the unsupervised model to [-1, 1] to ensure consistency with the training process. Formally, it can be defined as:
\begin{equation}
\begin{aligned}
x_u&=Unsuper(f), \\
x_T&=Scale(x_u),
\label{unsuper}
\end{aligned}
\end{equation}
where $Unsuper(\cdot)$ represents the unsupervised model, $x_u$ represents its generated importance scores, and $x_T$ represents the importance scores after scaling.

To adapt to the introduction of the unsupervised model, we reduce the maximum noise addition steps, allowing DDPM to start denoising from a relatively clear intermediate result instead of a Gaussian noise. Specifically, we treat the noising result $x_t$ in Eq.~\ref{predict} as a weighting of importance scores and noise, where $\sqrt{\bar{\alpha}_t}$ is the weight of importance scores and $\sqrt{1-\bar{\alpha}_t}$ is the weight of the noise. When the maximum noise addition steps set to 1000 and the variance schedule $\beta_1,..., \beta_t$ linearly increasing from $10^{-4}$ to $0.02$~\cite{ho2020denoising}, the value of $\sqrt{\bar{\alpha}_t}$ is close to 0, while the value of $\sqrt{1-\bar{\alpha}_t}$ is close to 1, thus the noising result is approximate to Gaussian noise. We adjust the weight of importance scores and noise by reducing the maximum noise addition steps so that the value of $\sqrt{\bar{\alpha}_t}$ gradually increases, while the value of $\sqrt{1-\bar{\alpha}_t}$ gradually decreases, thus achieving a relatively clear intermediate noising result.

A single denoising process can be defined as follows:
\begin{equation}
\begin{aligned}
x_{t-1}&=\frac{1}{\sqrt{\alpha_t}}(x_t-\frac{1-\alpha_t}{\sqrt{1-\bar{\alpha_t}}}\epsilon_\theta(x_t, f, t))+\sigma_tz, \\
\sigma_t&=\sqrt{\frac{1-\bar{\alpha}_{t-1}}{1-\bar{\alpha}_t}\beta_t},
\label{denoising}
\end{aligned}
\end{equation}
where $z$ is the random noise sampled from a standard normal distribution, and the complete testing process is shown in Algorithm 2. For each video, we use the output of the unsupervised model as the starting point and generate more accurate importance scores through gradual denoising.
\begin{algorithm}[tb]
\caption{Testing Process}
\textbf{Input:} Video features. \\
\textbf{Output:} Video summary.
\begin{algorithmic}
\FOR {$n\in\{1, 2, ..., N\}$}
    \STATE Extract video features $f$ of the $n$-th video.
    \STATE Generate importance scores $x_T$ using Eq.~\ref{unsuper}.
    \FOR {$t\in\{T, ..., 1\}$}
        \STATE $z\sim \mathcal{N}(0, 1)$ if $t>1$, else $z=0$.
        \STATE Denoise using Eq.~\ref{denoising}.
    \ENDFOR 
    \STATE Generate video summary through importance scores $x_0$.
\ENDFOR
\end{algorithmic}
\end{algorithm}

\begin{table*}[t]
\centering
\begin{tabular}{l|ccc|ccc}
     \toprule
     \multicolumn{1}{c|}{\multirow{2}{*}{Method}} & \multicolumn{3}{c|}{TVSum} & \multicolumn{3}{c}{SumMe} \\
     \cline{2-7}
     \multirow{2}{*}{} & Can & Aug & Tran & Can & Aug & Tran \\
     \hline
     dppLSTM~\cite{zhang2016video} & 54.7 & 59.6 & 58.7 & 38.6 & 42.9 & 41.8 \\
     VASNet~\cite{fajtl2019summarizing} & 61.4 & 62.4 & - & 49.7 & 51.1 & - \\
     SUM-FCN~\cite{rochan2018video} & 56.8 & 59.2 & 58.2 & 47.5 & 51.1 & 44.1 \\
     A-AVS~\cite{ji2019video} & 59.4 & 60.8 & - & 43.9 & 44.6 & - \\
     M-AVS~\cite{ji2019video} & 61.0 & 61.8 & - & 44.4 & 46.1 & - \\
     DSNet$_{ab}$~\cite{zhu2020dsnet} & 62.1 & 63.9 & 59.4 & 50.2 & 50.7 & 46.5 \\
     DSNet$_{af}$~\cite{zhu2020dsnet} & 61.9 & 62.2 & 58.0 & 51.2 & 53.3 & 47.6 \\
     RSGN~\cite{zhao2021reconstructive} & 60.1 & 61.1 & 60.0 & 45.0 & 45.7 & 44.0 \\
     3DST-UNet$_{sup}$~\cite{liu2022video} & 58.3 & 58.9 & 56.1 & 47.4 & 49.9 & 47.9 \\
     RR-STG~\cite{zhu2022relational} & 63.0 & 63.6 & 59.7 & 53.4 & 54.8 & 45.4 \\
     CFT-GIB$_{sup}$~\cite{zhong2023semantic} & 62.7 & 60.3 & 58.0 & 56.0 & 54.8 & 44.0 \\
     VSS-Net~\cite{zhang2023vss} & 61.0 & 61.4 & 58.5 & 51.5 & 52.8 & 48.4 \\
     LMVS~\cite{nam2024does} & 60.5 & - & - & 45.8 & - & - \\
     MPFN~\cite{khan2024deep} & 62.4 & - & - & 51.9 & - & - \\
     \hline
     SUM-GAN~\cite{mahasseni2017unsupervised} & 50.8 & 58.9 & - & 38.7 & 41.7 & - \\
     DR-DSN~\cite{zhou2018deep} & 57.6 & 58.4 & 57.8 & 41.4 & 42.8 & 42.4 \\
     ACGAN~\cite{he2019unsupervised} & 58.5 & 58.9 & 57.8 & 46.0 & 47.0 & 44.5 \\
     SUM-GAN-AAE~\cite{apostolidis2020unsupervised} & 58.3 & - & - & 48.9 & - & - \\
     3DST-UNet$_{unsup}$~\cite{liu2022video} & 58.3 & 58.4 & 58.0 & 44.6 & 49.5 & 45.7 \\
     CFT-GIB$_{unsup}$~\cite{zhong2023semantic} & 62.5 & 60.4 & 57.4 & 55.0 & 54.0 & 42.9 \\
     SSPVS~\cite{li2023progressive} & 60.3 & 61.8 & 57.8 & 48.7 & 50.4 & 45.8 \\
     AMFM~\cite{zhang2024attention} & 61.0 & 60.8 & 58.6 & 51.8 & 52.8 & 46.4 \\
     PRLVS~\cite{wang2024progressive} & 63.0 & 59.2 & 57.0 & 46.3 & 49.7 & 47.6 \\
     DMFF~\cite{yu2024unsupervised} & 61.2 & - & - & 53.0 & - & - \\
     \hline
     Ours & \textbf{64.8} & \textbf{65.0} & \textbf{60.9} & \textbf{58.7} & \textbf{59.2} & \textbf{50.3} \\
     \bottomrule
\end{tabular}
\caption{Performance comparison (F-score) with state-of-the-art video summarization methods on the TVSum and SumMe datasets under the canonical (Can), augmented (Aug), and transfer (Tran) settings.}
\label{Table2}
\end{table*}

\begin{table}[t]
\centering
\begin{tabular}{l|c}
     \toprule
     \multicolumn{1}{c|}{Method} & F-score \\
     \hline
     Random~\cite{ho2018summarizing}  & 16.3 \\
     Uniform~\cite{ho2018summarizing} & 15.1 \\
     TDCNN~\cite{yao2016highlight} & 28.6 \\
     DSN~\cite{bousmalis2016domain} & 22.7 \\
     FPVS~\cite{ho2018summarizing} & 35.3 \\
     \hline
     Ours & \textbf{46.1} \\
     \bottomrule
\end{tabular}
\caption{Performance comparison (F-score) with state-of-the-art video summarization methods on the FPVSum dataset.}
\label{Table3}
\end{table}

\begin{table}[t]
\centering
\begin{tabular}{l|cc|cc}
     \toprule
     \multicolumn{1}{c|}{\multirow{2}{*}{Method}} & \multicolumn{2}{c|}{TVSum} & \multicolumn{2}{c}{SumMe} \\
     \cline{2-5}
     \multirow{2}{*}{} & $\tau$ & $\rho$ & $\tau$ & $\rho$ \\
     \hline
     dppLSTM & 0.042 & 0.055 & 0.071 & 0.101 \\
     HSA & 0.082 & 0.088 & 0.064 & 0.066 \\
     DSNet$_{ab}$ & 0.108 & 0.129 & 0.051 & 0.059 \\
     DSNet$_{af}$ & 0.113 & 0.138 & 0.037 & 0.046 \\
     RSGN & 0.083 & 0.090 & 0.083 & 0.085 \\
     SSPVS & 0.177 & 0.233 & 0.178 & 0.240 \\
     \hline
     Ours & \textbf{0.179} & \textbf{0.238} & \textbf{0.221} & \textbf{0.252} \\
     \bottomrule
\end{tabular}
\caption{Performance comparison (Kendall’s $\tau$ and Spearman’s $\rho$ correlation coefficients) with state-of-the-art video summarization methods on the TVSum and SumMe datasets.}
\label{Table_c}
\end{table}


\section{Experiment}
\subsection{Dataset}
We conduct experiments on three benchmark datasets: TVSum~\cite{song2015tvsum}, SumMe~\cite{gong2014diverse} and FPVSum~\cite{ho2018summarizing}, in which TVSum consists of 50 videos, SumMe consists of 25 videos, and FPVSum consists of 56 labeled videos and 42 unlabeled videos. 
Following the protocol in~\cite{zhang2016video}, we build three settings for TVSum and SumMe: canonical, augmented and transfer. 
"canonical" is the standard supervised learning setting that divides the dataset into a training set and a testing set. "augmented" additionally introduces YouTube dataset~\cite{de2011vsumm} and Open Video Project (OVP) dataset~\cite{de2011vsumm} to explore the impact of the increased amount of annotations on performance. "transfer" divides completely different datasets into a training set and a testing set to simulate the cross-domain scenarios. Following the protocol in~\cite{ho2018summarizing}, we build the FPVSum setting as a supplement to the transfer setting. We divide videos with different points of view, introducing the third-person videos in TVSum and SumMe into the training set, and the first-person videos into the testing set. The details of dataset settings are shown in the appendix.
We perform validation experiments with 5 randomly created data splits and report the average results.

\subsection{Implementation Detail}
The training and testing processes are implemented using Pytorch. In training, the maximum step of noise addition is set to 200, and the ground-truth of importance scores are normalized to the range of -1 to 1 before adding Gaussian noise. In testing, the denoising step is set to 200, and the inverse process of training is conducted, which scales the generated importance scores to the range of 0 to 1. In addition, we use Adam as the optimizer and set the learning rate to 0.0002 and the weight decay to 0.01. The model is trained in 100 epochs, and a warmup strategy is used in the first 10 epochs.

\subsection{Quantitative Evaluation}
\subsubsection{ Comparison Results with the State-of-the-art Methods}
We compare our method with several state-of-the-art methods under different settings, including supervised methods (dppLSTM~\cite{zhang2016video}, VASNet~\cite{fajtl2019summarizing}, SUM-FCN~\cite{rochan2018video}, A-AVS~\cite{ji2019video}, M-AVS~\cite{ji2019video}, DSNet~\cite{zhu2020dsnet}, RSGN~\cite{zhao2021reconstructive}, 3DST-UNet$_{sup}$~\cite{liu2022video}, RR-STG~\cite{zhu2022relational}, CFT-GIB$_{sup}$~\cite{zhong2023semantic}, VSS-Net~\cite{zhang2023vss}, LMVS~\cite{nam2024does}, MPFN~\cite{khan2024deep}, TDCNN~\cite{yao2016highlight}, DSN~\cite{bousmalis2016domain} and FPVS~\cite{ho2018summarizing}), and unsupervised methods (SUM-GAN~\cite{mahasseni2017unsupervised}, DR-DSN~\cite{zhou2018deep}, ACGAN~\cite{he2019unsupervised}, SUM-GAN-AAE~\cite{apostolidis2020unsupervised}, 3DST-UNet$_{unsup}$~\cite{liu2022video}, CFT-GIB$_{unsup}$~\cite{zhong2023semantic}, SSPVS~\cite{li2023progressive}, AMFM~\cite{zhang2024attention}, PRLVS~\cite{wang2024progressive}, DMFF~\cite{yu2024unsupervised} Random~\cite{ho2018summarizing} and Uniform~\cite{ho2018summarizing}).

Table~\ref{Table2} and~\ref{Table3} show the comparison results on the TVSum, SumMe, and FPVSum datasets. It is interesting to observe that our method performs best under all the settings on all the datasets, which indicates that compared with the discriminative methods, our generative method is more resistant to subjective annotation noise. In addition, our method achieves much better results with large gains on the transfer settings on SumMe and TVSum as well as FPVSum dataset, which suggests that our method is less prone to overfitting the training data and has stronger generalization ability.

Besides the F-Score, Kendall’s $\tau$ and Spearman’s $\rho$ correlation coefficients are applied in~\cite{otani2019rethinking} as alternatives for assessing the importance score. Table~\ref{Table_c} shows the comparison results of our method with the state-of-the-art methods using Kendall’s $\tau$ and Spearman’s $\rho$ correlation coefficients on TVSum and SumMe, including dppLSTM~\cite{zhang2016video}, HSA~\cite{zhao2018hsa}, DSNet$_{ab}$~\cite{zhu2020dsnet}, DSNet$_{af}$~\cite{zhu2020dsnet}, RSGN~\cite{zhao2021reconstructive}, SSPVS~\cite{li2023progressive}. 
We observe that the ranks of the proposed method also achieve the best result, which further demonstrates the effectiveness of our method.

\begin{table}[t]
\setlength{\tabcolsep}{1mm}
\centering
\begin{tabular}{l|ccc|ccc|c}
     \toprule
     \multicolumn{1}{c|}{\multirow{2}{*}{Method}} & \multicolumn{3}{c|}{TVSum} & \multicolumn{3}{c|}{SumMe} & \multirow{2}{*}{FVPSum} \\
     \cline{2-7}
     \multirow{2}{*}{} & Can & Aug & Tran & Can & Aug & Tran & \multirow{2}{*}{} \\
     \hline
     w/o DDPM & 62.0 & 61.5 & 55.3 & 51.1 & 51.1 & 46.5 & 39.1 \\
     w/o unsup & 64.1 & 62.7 & 57.6 & 51.9 & 57.5 & 48.1 & 42.3 \\ 
     \hline
     Ours & \textbf{64.8} & \textbf{65.0} & \textbf{60.9} & \textbf{58.7} & \textbf{59.2} & \textbf{50.3} & \textbf{46.1} \\
     \bottomrule
\end{tabular}
\caption{Ablation study results (F-score) on the TVSum, SumMe, and FPVSum datasets.}
\label{Table4}
\end{table}


\subsubsection{Ablation Study}
To perform an in-depth analysis of each individual component of our method, we conduct extensive ablation studies on TVSum, SumMe, and FPVSum. We employ variants of our method for comparison, including ``w/o DDPM" and ``w/o unsup". ``w/o DDPM"  represents removing DDPM and using only the unsupervised video summarization model, in order to evaluate the effectiveness of DDPM. ``w/o unsup" represents removing the unsupervised video summarization model and using only DDPM, in order to evaluate the effectiveness of unsupervised video summarization model, which is achieved by setting the maximum noise addition steps to 1000 during training, and using Gaussian noise as the starting point for the denoising process during testing.
From the results in Table~\ref{Table4}, we observe that our method achieves better performance than ``w/o DDPM", indicating that DDPM succeeds in generating high-quality summaries by the denoising process. Our method also has advantages over ``w/o unsup", indicating that using an unsupervised method to replace the early denoising process of DDPM is an effective strategy under data scarcity.

In addition, we evaluate the impact of different unsupervised models (SUM-GAN-AAE~\cite{apostolidis2020unsupervised}, AC-SUM-GAN~\cite{apostolidis2020ac}, and DR-DSN~\cite{zhou2018deep}) on TVSum and SumMe. The results are shown in Table~\ref{Table5} and we choose the DR-DSN with the best overall performance as the unsupervised model for our method. We observe that in some settings, different unsupervised models exhibit similar performance, indicating that our method is not limited to a specific unsupervised model and has robustness to different unsupervised models.

\begin{figure}[t]
  \centering
  \includegraphics[width=\linewidth]{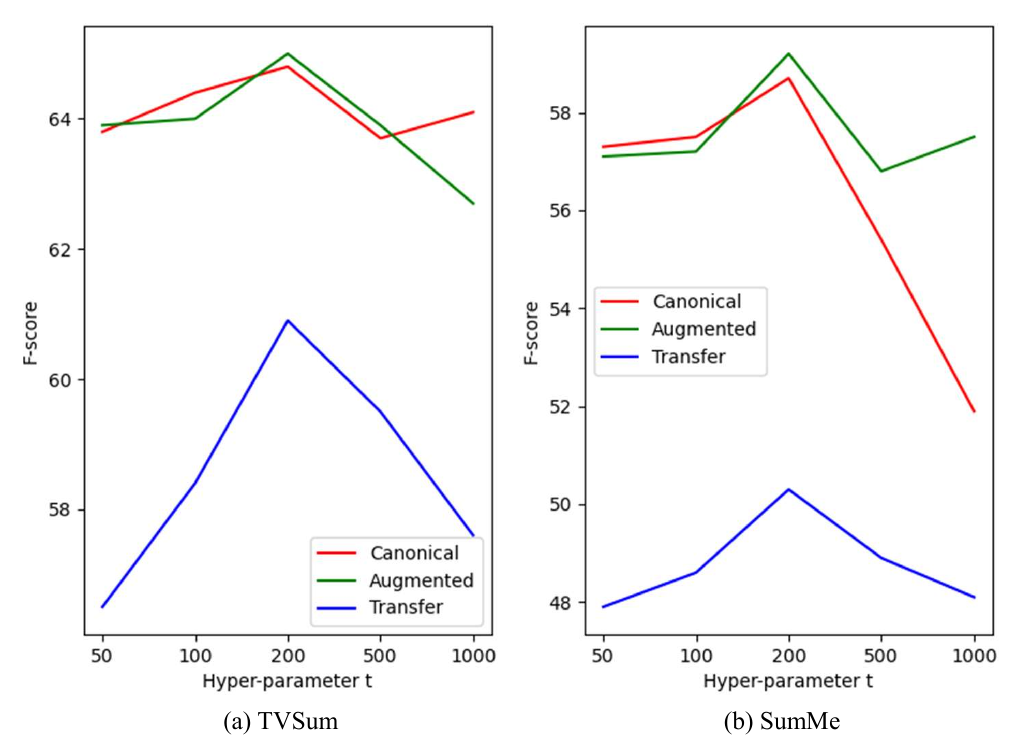}
  \caption{Results (F-score) of experiment with different hyper-parameter $t$ on the TVSum and SumMe datasets.}
  \label{Figure4}
\end{figure}

\begin{table}[t]
\setlength{\tabcolsep}{1mm}
\centering
\begin{tabular}{l|ccc|ccc}
     \toprule
     \multicolumn{1}{c|}{\multirow{2}{*}{Method}} & \multicolumn{3}{c|}{TVSum} & \multicolumn{3}{c}{SumMe} \\
     \cline{2-7}
     \multirow{2}{*}{} & Can & Aug & Tran & Can & Aug & Tran \\
     \hline
     SUM-GAN-AAE & 64.1 & 63.9 & 58.4 & 56.9 & 58.1 & 47.6 \\
     AC-SUM-GAN & 63.5 & \textbf{65.3} & 58.5 & 56.1 & 58.5 & 48.8 \\ 
     DR-DSN & \textbf{64.8} & 65.0 & \textbf{60.9} & \textbf{58.7} & \textbf{59.2} & \textbf{50.3} \\
     \bottomrule
\end{tabular}
\caption{Results (F-score) of experiment with different unsupervised methods on the TVSum and SumMe datasets.}
\label{Table5}
\end{table}

\begin{figure*}[t]
  \centering
  \includegraphics[width=1.02\textwidth]{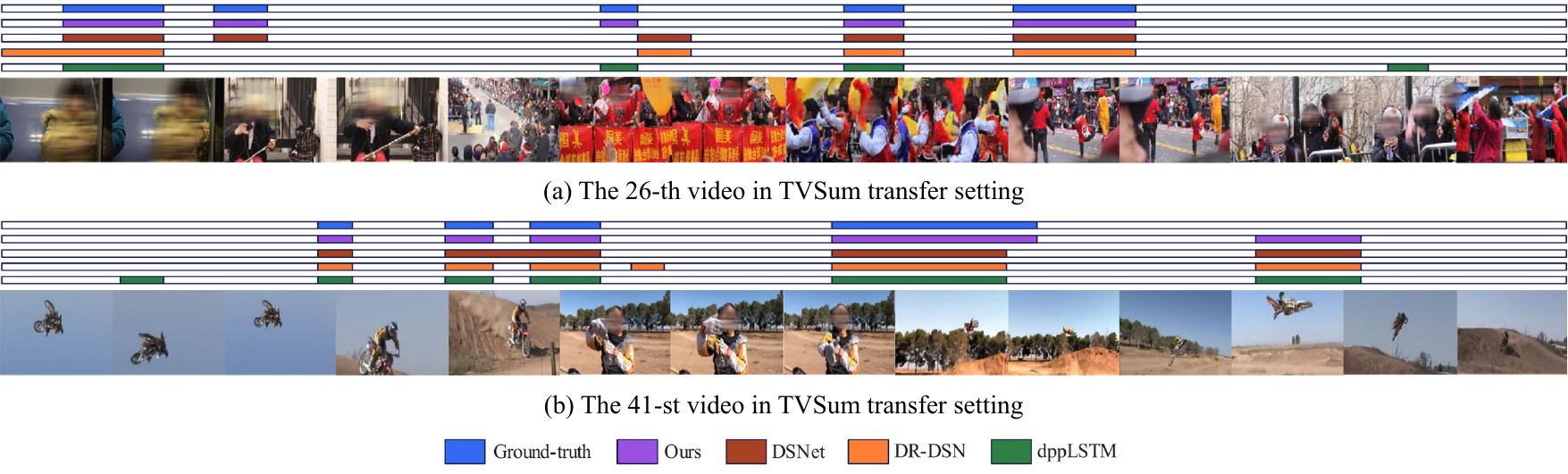}
  \caption{Qualitative results of different video summarization methods. The line segments denote the selected segments and the frames are shown below.}
  \label{Figure5}
\end{figure*}

\begin{figure}[t]
  \centering
  \includegraphics[width=0.82\linewidth]{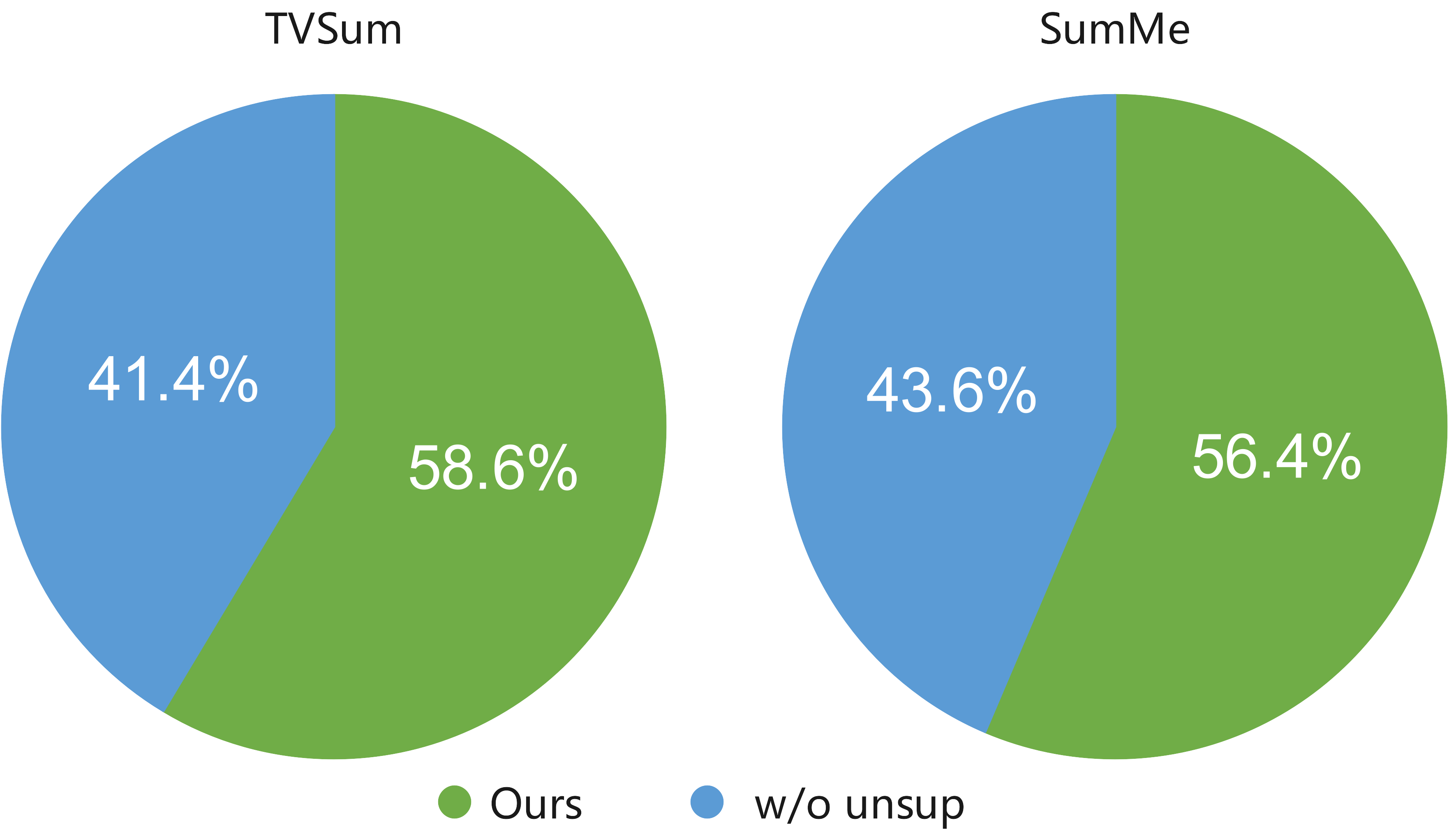}
  \caption{Human evaluation of video summaries generated by 'w/o unsup' and our method. The blue sector represents the percentage of summaries chosen by users from 'w/o unsup' and the green sector represents the percentage of summaries chosen by users from our method.}
  \label{Figure7}
\end{figure}

\subsubsection{Analysis of Hyper-parameter}
To analyze the denoising ability of DDPM in video summarization, we evaluate the impact of hyper-parameter $t$ on TVSum and SumMe. The results are shown in Figure~\ref{Figure4}. When t=1000, DDPM denoises from Gaussian noise without unsupervised model. It is interesting to observe that for most settings, our method achieves the best performance when $t$ equals 200. This indicates that the importance scores generated by the unsupervised video summarization model can be approximated as the ground-truth with 200 steps of noise addition, and DDPM can generate better quality summaries on this basis by 200 steps of denoising. But, when $t$ equals to 50 or 1000, both the performance significantly degrades. The former shows that the importance scores generated by the unsupervised video summarization model are not accurate enough and it is infeasible to obtain a satisfactory summary using only 50 steps of denoising, while the latter shows that the denoising ability of DDPM is not sufficient to directly generate summaries from completely Gaussian noise.

\subsection{Qualitative Evaluation}
\subsubsection{Qualitative Comparison Results}
Figure~\ref{Figure5} demonstrates several examples of generated video summaries by DSNet~\cite{zhu2020dsnet}, DR-DSN~\cite{zhou2018deep},  dppLSTM~\cite{zhang2016video}, our method and ground-truth. We can observe that in the cases, the summaries generated by our method have more overlaps with the ground-truth summaries, which demonstrates that our method effectively reduces the interference of subjective annotation noise and generates high-quality summaries. In the cross-domain scenarios shown in Figure~\ref{Figure5} (b), our method outperforms other methods, which exhibits stronger generalization ability. More experiments of the visualization cases are shown in the appendix.

\subsubsection{User Study}
We also conduct human evaluation to analyze the effectiveness of our method. Concretely, for all videos in TVSum and SumMe datasets, we show each user the video summaries generated by ``w/o unsup" and the video summaries generated by our method. 10 users are asked to choose the video summaries they are more interested in.
Figure~\ref{Figure7} shows the human evaluation results. For each dataset, the blue sector represents the percentage of summaries generated by ``w/o unsup" that are chosen by users and the green sector represents the percentage of summaries generated by our complete method that are chosen by users. We observe that the video summaries generated by our method are more preferred by users for both datasets, which further indicates that our method has advantages over removing the unsupervised video summarization model and using only DDPM.

\section{Conclusion}
We have presented a novel diffusion method for video summarization based on Denoising Diffusion Probabilistic Model (DDPM). It learns the probability distribution of training data through noise prediction, and generates summaries by iterative denoising. Our method can resist subjective annotation noise with better robustness and less overfit the training data with stronger generalization, compared to the discriminative method. Extensive experiments on different settings of various datasets demonstrate the effectiveness of our method. 

In the future, we are going to apply our method to query-based video summarization, introducing user preferences into the summarization process in the form of text queries, and further exploring the impact of user subjectivity on video summarization.

\bibliography{aaai25}

\begin{thebibliography}{43}
\providecommand{\natexlab}[1]{#1}

\bibitem[{Apostolidis et~al.(2020{\natexlab{a}})Apostolidis, Adamantidou, Metsai, Mezaris, and Patras}]{apostolidis2020ac}
Apostolidis, E.; Adamantidou, E.; Metsai, A.~I.; Mezaris, V.; and Patras, I. 2020{\natexlab{a}}.
\newblock AC-SUM-GAN: Connecting actor-critic and generative adversarial networks for unsupervised video summarization.
\newblock \emph{IEEE Transactions on Circuits and Systems for Video Technology}, 31(8): 3278--3292.

\bibitem[{Apostolidis et~al.(2020{\natexlab{b}})Apostolidis, Adamantidou, Metsai, Mezaris, and Patras}]{apostolidis2020unsupervised}
Apostolidis, E.; Adamantidou, E.; Metsai, A.~I.; Mezaris, V.; and Patras, I. 2020{\natexlab{b}}.
\newblock Unsupervised video summarization via attention-driven adversarial learning.
\newblock In \emph{Proceedings of the International Conference on MultiMedia Modeling}, 492--504. Springer.

\bibitem[{Bousmalis et~al.(2016)Bousmalis, Trigeorgis, Silberman, Krishnan, and Erhan}]{bousmalis2016domain}
Bousmalis, K.; Trigeorgis, G.; Silberman, N.; Krishnan, D.; and Erhan, D. 2016.
\newblock Domain separation networks.
\newblock \emph{Advances in Neural Information Processing Systems}, 29.

\bibitem[{Carrillo et~al.(2023)Carrillo, Cl{\'e}ment, Bugeau, and Simo-Serra}]{carrillo2023diffusart}
Carrillo, H.; Cl{\'e}ment, M.; Bugeau, A.; and Simo-Serra, E. 2023.
\newblock Diffusart: Enhancing line art colorization with conditional diffusion models.
\newblock In \emph{Proceedings of the IEEE/CVF Conference on Computer Vision and Pattern Recognition}, 3486--3490.

\bibitem[{De~Avila et~al.(2011)De~Avila, Lopes, da~Luz~Jr, and de~Albuquerque~Ara{\'u}jo}]{de2011vsumm}
De~Avila, S. E.~F.; Lopes, A. P.~B.; da~Luz~Jr, A.; and de~Albuquerque~Ara{\'u}jo, A. 2011.
\newblock VSUMM: A mechanism designed to produce static video summaries and a novel evaluation method.
\newblock \emph{Pattern Recognition Letters}, 32(1): 56--68.

\bibitem[{Fajtl et~al.(2019)Fajtl, Sokeh, Argyriou, Monekosso, and Remagnino}]{fajtl2019summarizing}
Fajtl, J.; Sokeh, H.~S.; Argyriou, V.; Monekosso, D.; and Remagnino, P. 2019.
\newblock Summarizing videos with attention.
\newblock In \emph{Proceedings of the Asian Conference on Computer Vision}, 39--54. Springer.

\bibitem[{Gao et~al.(2023)Gao, Liu, Zeng, Xu, Li, Luo, Liu, Zhen, and Zhang}]{gao2023implicit}
Gao, S.; Liu, X.; Zeng, B.; Xu, S.; Li, Y.; Luo, X.; Liu, J.; Zhen, X.; and Zhang, B. 2023.
\newblock Implicit diffusion models for continuous super-resolution.
\newblock In \emph{Proceedings of the IEEE/CVF Conference on Computer Vision and Pattern Recognition}, 10021--10030.

\bibitem[{Gong et~al.(2014)Gong, Chao, Grauman, and Sha}]{gong2014diverse}
Gong, B.; Chao, W.-L.; Grauman, K.; and Sha, F. 2014.
\newblock Diverse sequential subset selection for supervised video summarization.
\newblock \emph{Advances in Neural Information Processing Systems}, 27.

\bibitem[{He et~al.(2019)He, Hua, Song, Zhang, Xue, Ma, Robertson, and Guan}]{he2019unsupervised}
He, X.; Hua, Y.; Song, T.; Zhang, Z.; Xue, Z.; Ma, R.; Robertson, N.; and Guan, H. 2019.
\newblock Unsupervised video summarization with attentive conditional generative adversarial networks.
\newblock In \emph{Proceedings of the 27th ACM International Conference on Multimedia}, 2296--2304.

\bibitem[{Ho, Chiu, and Wang(2018)}]{ho2018summarizing}
Ho, H.-I.; Chiu, W.-C.; and Wang, Y.-C.~F. 2018.
\newblock Summarizing first-person videos from third persons' points of view.
\newblock In \emph{Proceedings of the European Conference on Computer Vision}, 70--85.

\bibitem[{Ho, Jain, and Abbeel(2020)}]{ho2020denoising}
Ho, J.; Jain, A.; and Abbeel, P. 2020.
\newblock Denoising diffusion probabilistic models.
\newblock \emph{Advances in Neural Information Processing Systems}, 33: 6840--6851.

\bibitem[{Hsu, Liao, and Huang(2023)}]{hsu2023video}
Hsu, T.-C.; Liao, Y.-S.; and Huang, C.-R. 2023.
\newblock Video summarization with spatiotemporal vision transformer.
\newblock \emph{IEEE Transactions on Image Processing}, 32: 3013--3026.

\bibitem[{Ji et~al.(2019)Ji, Xiong, Pang, and Li}]{ji2019video}
Ji, Z.; Xiong, K.; Pang, Y.; and Li, X. 2019.
\newblock Video summarization with attention-based encoder--decoder networks.
\newblock \emph{IEEE Transactions on Circuits and Systems for Video Technology}, 30(6): 1709--1717.

\bibitem[{Khan et~al.(2024)Khan, Hussain, Khan, Khan, and Baik}]{khan2024deep}
Khan, H.; Hussain, T.; Khan, S.~U.; Khan, Z.~A.; and Baik, S.~W. 2024.
\newblock Deep multi-scale pyramidal features network for supervised video summarization.
\newblock \emph{Expert Systems with Applications}, 237: 121288.

\bibitem[{Lagani{\`e}re et~al.(2008)Lagani{\`e}re, Bacco, Hocevar, Lambert, Pa{\"\i}s, and Ionescu}]{laganiere2008video}
Lagani{\`e}re, R.; Bacco, R.; Hocevar, A.; Lambert, P.; Pa{\"\i}s, G.; and Ionescu, B.~E. 2008.
\newblock Video summarization from spatio-temporal features.
\newblock In \emph{Proceedings of the 2nd ACM TRECVid Video Summarization Workshop}, 144--148.

\bibitem[{Li et~al.(2023)Li, Ke, Gong, and Drummond}]{li2023progressive}
Li, H.; Ke, Q.; Gong, M.; and Drummond, T. 2023.
\newblock Progressive video summarization via multimodal self-supervised learning.
\newblock In \emph{Proceedings of the IEEE/CVF Winter Conference on Applications of Computer Vision}, 5584--5593.

\bibitem[{Li et~al.(2022)Li, Yang, Chang, Chen, Feng, Xu, Li, and Chen}]{li2022srdiff}
Li, H.; Yang, Y.; Chang, M.; Chen, S.; Feng, H.; Xu, Z.; Li, Q.; and Chen, Y. 2022.
\newblock Srdiff: Single image super-resolution with diffusion probabilistic models.
\newblock \emph{Neurocomputing}, 479: 47--59.

\bibitem[{Liang et~al.(2022)Liang, Lv, Li, Zhang, and Zhang}]{liang2022video}
Liang, G.; Lv, Y.; Li, S.; Zhang, S.; and Zhang, Y. 2022.
\newblock Video summarization with a convolutional attentive adversarial network.
\newblock \emph{Pattern Recognition}, 131: 108840.

\bibitem[{Liu et~al.(2022)Liu, Meng, Huang, Vlontzos, Rueckert, and Kainz}]{liu2022video}
Liu, T.; Meng, Q.; Huang, J.-J.; Vlontzos, A.; Rueckert, D.; and Kainz, B. 2022.
\newblock Video summarization through reinforcement learning with a 3D spatio-temporal u-net.
\newblock \emph{IEEE Transactions on Image Processing}, 31: 1573--1586.

\bibitem[{Mahasseni, Lam, and Todorovic(2017)}]{mahasseni2017unsupervised}
Mahasseni, B.; Lam, M.; and Todorovic, S. 2017.
\newblock Unsupervised video summarization with adversarial lstm networks.
\newblock In \emph{Proceedings of the IEEE conference on Computer Vision and Pattern Recognition}, 202--211.

\bibitem[{Nam et~al.(2024)Nam, Lehavi, Yang, Bose, Swayamdipta, and Narayanan}]{nam2024does}
Nam, Y.; Lehavi, A.; Yang, D.; Bose, D.; Swayamdipta, S.; and Narayanan, S. 2024.
\newblock Does Video Summarization Require Videos? Quantifying the Effectiveness of Language in Video Summarization.
\newblock In \emph{Proceedings of the International Conference on Acoustics, Speech and Signal Processing}, 8396--8400. IEEE.

\bibitem[{Nichol and Dhariwal(2021)}]{nichol2021improved}
Nichol, A.~Q.; and Dhariwal, P. 2021.
\newblock Improved denoising diffusion probabilistic models.
\newblock In \emph{Proceedings of the International Conference on Machine Learning}, 8162--8171. PMLR.

\bibitem[{Otani et~al.(2019)Otani, Nakashima, Rahtu, and Heikkila}]{otani2019rethinking}
Otani, M.; Nakashima, Y.; Rahtu, E.; and Heikkila, J. 2019.
\newblock Rethinking the evaluation of video summaries.
\newblock In \emph{Proceedings of the IEEE/CVF Conference on Computer Vision and Pattern Recognition}, 7596--7604.

\bibitem[{Ren et~al.(2017)Ren, Yan, Ni, Liu, Yang, and Zha}]{ren2017unsupervised}
Ren, Z.; Yan, J.; Ni, B.; Liu, B.; Yang, X.; and Zha, H. 2017.
\newblock Unsupervised deep learning for optical flow estimation.
\newblock In \emph{Proceedings of the AAAI Conference on Artificial Intelligence}, volume~31.

\bibitem[{Rochan, Ye, and Wang(2018)}]{rochan2018video}
Rochan, M.; Ye, L.; and Wang, Y. 2018.
\newblock Video summarization using fully convolutional sequence networks.
\newblock In \emph{Proceedings of the European Conference on Computer Vision}, 347--363.

\bibitem[{Ronneberger, Fischer, and Brox(2015)}]{ronneberger2015u}
Ronneberger, O.; Fischer, P.; and Brox, T. 2015.
\newblock U-net: Convolutional networks for biomedical image segmentation.
\newblock In \emph{Proceedings of the International Conference on Medical Image Computing and Computer-Assisted Intervention}, 234--241. Springer.

\bibitem[{Singh and Kumar(2024)}]{singh2024bayesian}
Singh, A.; and Kumar, M. 2024.
\newblock Bayesian fuzzy clustering and deep CNN-based automatic video summarization.
\newblock \emph{Multimedia Tools and Applications}, 83(1): 963--1000.

\bibitem[{Song et~al.(2015)Song, Vallmitjana, Stent, and Jaimes}]{song2015tvsum}
Song, Y.; Vallmitjana, J.; Stent, A.; and Jaimes, A. 2015.
\newblock Tvsum: Summarizing web videos using titles.
\newblock In \emph{Proceedings of the IEEE/CVF Conference on Computer Vision and Pattern Recognition}, 5179--5187.

\bibitem[{Terbouche et~al.(2023)Terbouche, Morel, Rodriguez, and Othmani}]{terbouche2023multi}
Terbouche, H.; Morel, M.; Rodriguez, M.; and Othmani, A. 2023.
\newblock Multi-annotation attention model for video summarization.
\newblock In \emph{Proceedings of the IEEE/CVF Conference on Computer Vision and Pattern Recognition}, 3143--3152.

\bibitem[{Wang, Wu, and Yan(2024)}]{wang2024progressive}
Wang, G.; Wu, X.; and Yan, J. 2024.
\newblock Progressive reinforcement learning for video summarization.
\newblock \emph{Information Sciences}, 655: 119888.

\bibitem[{Wu and De~la Torre(2023)}]{wu2023latent}
Wu, C.~H.; and De~la Torre, F. 2023.
\newblock A latent space of stochastic diffusion models for zero-shot image editing and guidance.
\newblock In \emph{Proceedings of the IEEE/CVF International Conference on Computer Vision}, 7378--7387.

\bibitem[{Yao, Mei, and Rui(2016)}]{yao2016highlight}
Yao, T.; Mei, T.; and Rui, Y. 2016.
\newblock Highlight detection with pairwise deep ranking for first-person video summarization.
\newblock In \emph{Proceedings of the IEEE/CVF Conference on Computer Vision and Pattern Recognition}, 982--990.

\bibitem[{Yu et~al.(2024)Yu, Yu, Sun, Ding, and Jian}]{yu2024unsupervised}
Yu, Q.; Yu, H.; Sun, Y.; Ding, D.; and Jian, M. 2024.
\newblock Unsupervised Video Summarization Based on the Diffusion Model of Feature Fusion.
\newblock \emph{IEEE Transactions on Computational Social Systems}.

\bibitem[{Zhang et~al.(2016)Zhang, Chao, Sha, and Grauman}]{zhang2016video}
Zhang, K.; Chao, W.-L.; Sha, F.; and Grauman, K. 2016.
\newblock Video summarization with long short-term memory.
\newblock In \emph{Proceedings of the European Conference on Computer Vision}, 766--782. Springer.

\bibitem[{Zhang et~al.(2023)Zhang, Liu, Kang, and Tao}]{zhang2023vss}
Zhang, Y.; Liu, Y.; Kang, W.; and Tao, R. 2023.
\newblock VSS-Net: visual semantic self-mining network for video summarization.
\newblock \emph{IEEE Transactions on Circuits and Systems for Video Technology}.

\bibitem[{Zhang, Liu, and Wu(2024)}]{zhang2024attention}
Zhang, Y.; Liu, Y.; and Wu, C. 2024.
\newblock Attention-guided multi-granularity fusion model for video summarization.
\newblock \emph{Expert Systems with Applications}, 249: 123568.

\bibitem[{Zhao et~al.(2021)Zhao, Li, Lu, and Li}]{zhao2021reconstructive}
Zhao, B.; Li, H.; Lu, X.; and Li, X. 2021.
\newblock Reconstructive sequence-graph network for video summarization.
\newblock \emph{IEEE Transactions on Pattern Analysis and Machine Intelligence}, 44(5): 2793--2801.

\bibitem[{Zhao, Li, and Lu(2017)}]{zhao2017hierarchical}
Zhao, B.; Li, X.; and Lu, X. 2017.
\newblock Hierarchical recurrent neural network for video summarization.
\newblock In \emph{Proceedings of the 25th ACM international conference on Multimedia}, 863--871.

\bibitem[{Zhao, Li, and Lu(2018)}]{zhao2018hsa}
Zhao, B.; Li, X.; and Lu, X. 2018.
\newblock Hsa-rnn: Hierarchical structure-adaptive rnn for video summarization.
\newblock In \emph{Proceedings of the IEEE/CVF Conference on Computer Vision and Pattern Recognition}, 7405--7414.

\bibitem[{Zhong et~al.(2023)Zhong, Wang, Yao, Hu, Dong, and Munteanu}]{zhong2023semantic}
Zhong, R.; Wang, R.; Yao, W.; Hu, M.; Dong, S.; and Munteanu, A. 2023.
\newblock Semantic representation and attention alignment for Graph Information Bottleneck in video summarization.
\newblock \emph{IEEE Transactions on Image Processing}.

\bibitem[{Zhou, Qiao, and Xiang(2018)}]{zhou2018deep}
Zhou, K.; Qiao, Y.; and Xiang, T. 2018.
\newblock Deep reinforcement learning for unsupervised video summarization with diversity-representativeness reward.
\newblock In \emph{Proceedings of the AAAI Conference on Artificial Intelligence (AAAI)}, volume~32.

\bibitem[{Zhu et~al.(2022)Zhu, Han, Lu, and Zhou}]{zhu2022relational}
Zhu, W.; Han, Y.; Lu, J.; and Zhou, J. 2022.
\newblock Relational reasoning over spatial-temporal graphs for video summarization.
\newblock \emph{IEEE Transactions on Image Processing}, 31: 3017--3031.

\bibitem[{Zhu et~al.(2020)Zhu, Lu, Li, and Zhou}]{zhu2020dsnet}
Zhu, W.; Lu, J.; Li, J.; and Zhou, J. 2020.
\newblock Dsnet: A flexible detect-to-summarize network for video summarization.
\newblock \emph{IEEE Transactions on Image Processing}, 30: 948--962.

\end{thebibliography}

\end{document}